\pdfoutput=1

\documentclass[11pt]{article}

\usepackage[preprint]{acl}

\usepackage{times}
\usepackage{latexsym}
\usepackage{float} 
\usepackage{graphicx} 
\usepackage{booktabs} 
\usepackage{makecell} 
\usepackage[T1]{fontenc}

\usepackage[utf8]{inputenc}

\usepackage{microtype}

\usepackage{inconsolata}
\usepackage{booktabs}
\usepackage{graphicx}
\usepackage{amsmath}
\usepackage{enumitem}
\usepackage{stfloats}

\title{MKG-Rank: Enhancing Large Language Models with Knowledge Graph\\ for Multilingual Medical Question Answering}

\author{
 \textbf{Feiyang Li\textsuperscript{1}},
 \textbf{Yingjian Chen\textsuperscript{1}},
 \textbf{Haoran Liu\textsuperscript{2}},
 \textbf{Rui Yang\textsuperscript{3}},
 \textbf{Han Yuan\textsuperscript{3}},
 \textbf{Yuang Jiang\textsuperscript{4}},
 \\
 \textbf{Tianxiao Li\textsuperscript{5}},
 \textbf{Edison Marrese Taylor\textsuperscript{1}},
 \textbf{Hossein Rouhizadeh\textsuperscript{6}},\\
 \textbf{Yusuke Iwasawa\textsuperscript{1}}, 
 \textbf{Douglas Teodoro\textsuperscript{6}},
 \textbf{Yutaka Matsuo\textsuperscript{1}},
 \textbf{Irene Li\textsuperscript{1}}\thanks{Corresponding author}
\\
\\
 \textsuperscript{1}University of Tokyo,
 \textsuperscript{2}Texas A\&M University,
  \textsuperscript{3}Duke-NUS Medical School,
 \\
 \textsuperscript{4}Smartor LLC, Japan,
 \textsuperscript{5}NEC Laboratories America,
\textsuperscript{6}University of Geneva, Switzerland
\\
}

\begin{document}

\maketitle
\begin{abstract}
Large Language Models (LLMs) have shown remarkable progress in medical question answering (QA), yet their effectiveness remains predominantly limited to English due to imbalanced multilingual training data and scarce medical resources for low-resource languages. To address this critical language gap in medical QA, we propose \textbf{M}ultilingual \textbf{K}nowledge \textbf{G}raph-based Retrieval \textbf{Rank}ing (\textbf{MKG-Rank}), a knowledge graph-enhanced framework that enables English-centric LLMs to perform multilingual medical QA. Through a word-level translation mechanism, our framework efficiently integrates comprehensive English-centric medical knowledge graphs into LLM reasoning at a low cost, mitigating cross-lingual semantic distortion and achieving precise medical QA across language barriers. To enhance efficiency, we introduce caching and multi-angle ranking strategies to optimize the retrieval process, significantly reducing response times and prioritizing relevant medical knowledge. Extensive evaluations on multilingual medical QA benchmarks across Chinese, Japanese, Korean, and Swahili demonstrate that MKG-Rank consistently outperforms zero-shot LLMs, achieving maximum 35.03\% increase in accuracy, while maintaining an average retrieval time of only 0.0009 seconds.\footnote{Our anonymous code is available at \url{https://anonymous.4open.science/r/MKG-Rank-6B72}.}
\end{abstract}

\section{Introduction}

Large Language Models (LLMs)~\cite{hurst2024gpt, claude35sonnet, dubey2024llama} have achieved remarkable performance in a wide range of Natural Language Processing (NLP) tasks, including question answering~\cite{jiang2021can, dong2022closed} and information retrieval~\cite{Wang2023KnowledgeDrivenCE, Fan2024ASO}. Beyond general NLP applications, LLMs have also been successfully applied to specific professional domains such as medicine and law, demonstrating promising results~\cite{yang2024kg,yang2024ascle,ke2024mitigating,zakka2024almanac,Bernsohn2024LegalLensLL}.


While these advances have been demonstrated predominantly in English-language settings, research on their effectiveness in other languages remains relatively underexplored, especially in medical question-answering \cite{Singh2024INDICQB}. 
Specifically, the remaining challenges are: (1) mainstream LLMs are predominantly trained with English-centric data, resulting in a highly unbalanced distribution between languages~\cite{chataigner2024multilingual}, which limits their effectiveness in multilingual contexts; and (2) high-quality external medical data for low-resource languages are extremely scarce~\cite{Quercia2024MedFrenchmarkAS}. As a result, existing LLMs exhibit significant performance gaps in multilingual medical applications, limiting their use in non-English-speaking medical settings.

\noindent\textbf{Existing Works and Limitations.} Existing methods have emerged but still face significant limitations. 
\textit{Translation-based methods} either translate inputs into English for inference~\cite{asai2018multilingual, montero-etal-2022-pivot} or convert rich English corpora into target languages to generate training data~\cite{jundi-lapesa-2022-translate, zhang2023chinese}, both of which incur high translation costs and introduce medical semantic distortion. 
Alternatively, \textit{data-intensive adaptation techniques}~\cite{yang2023bigtranslate, lai2023okapi, li2023bactrian, ustun2024aya} rely on massive multilingual corpora, which are scarce in specialized medical domains. 
While recent \textit{multilingual RAG systems}~\cite{chirkova-etal-2024-retrieval, yang2024language, park2025investigating} avoid the need for retraining, they still depend on external multilingual databases, which are scarce or unavailable in low-resource medical contexts.

\noindent\textbf{Our Approach.}
To address the mentioned challenges in multilingual medical QA, we propose \textbf{M}ultilingual \textbf{K}nowledge \textbf{G}raph-based Retrieval \textbf{Rank}ing (\textbf{MKG-Rank}), a knowledge graph-augmented framework that uses comprehensive and easily accessible external medical knowledge graphs, enabling English-centric LLMs to achieve multilingual medical QA at minimal cost. Additionally, we design caching and multi-angle ranking strategies to improve the efficiency of the retrieval process and filter out irrelevant information. 

In summary, our contributions are: (1) we propose MKG-Rank, an efficient framework that enables English-centric LLMs to handle multilingual medical QA by using easily accessible external medical knowledge graphs; (2) we introduce a word-level translation mechanism that ensures precise medical term translation at minimal cost while preventing semantic distortion; and (3) we conduct extensive experiments on four multilingual medical QA datasets, showing that MKG-Rank consistently outperforms zero-shot base LLMs, with accuracy improvements of up to 35.03\%.

\begin{figure*}[t]
    \vspace{-3mm}
    \centering
    \includegraphics[width=1\linewidth]{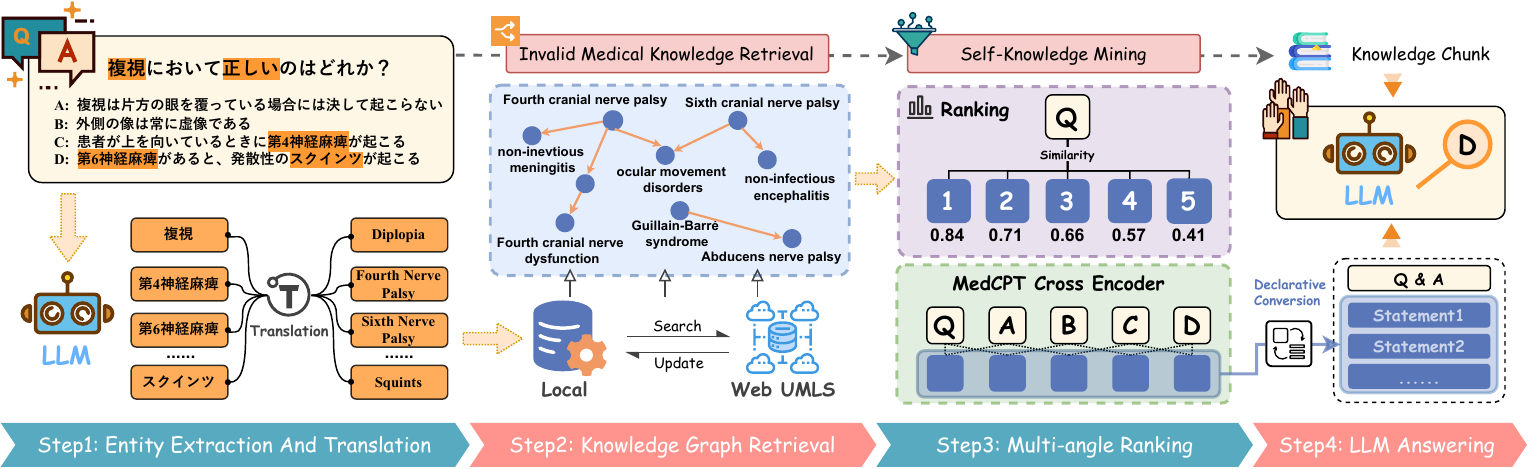}
    \caption{The overall architecture of our proposed MKG-Rank. The English translation of the question and options in the figure is provided in the Appendix~\ref{appendix_pipline}.}
    \label{fig:pipeline}
\end{figure*}

\section{Methodology}

In this section, we introduce our MKG-Rank, as illustrated in Figure~\ref{fig:pipeline}. The MKG-Rank framework includes four steps. First, we extract medical entities from the question and options and translate them into English. Then, translated medical entities are used to query an external medical knowledge base (UMLS~\cite{bodenreider2004unified}) and retrieve relevant knowledge graphs (KGs). Next, we propose a multi-angle ranking strategy to select the most relevant medical triplets. Finally, these triplets are first converted into declarative sentences and then, along with the original question and options, fed into the LLM for inference. Additionally, considering the case of invalid external medical knowledge retrieval, we perform self-knowledge mining to acquire relevant knowledge chunks, enhancing LLM's reasoning in medical QA.






\subsection{Medical Entity Extraction and Translation}  
Given a medical question $T_q$ and options $T_o$, we first use an LLM to extract relevant medical entities. Specifically, we extract up to three entities from the question and at most one entity from each option. The extracted entities are then translated into English, forming the set of medical entities used for retrieval, denoted as $\mathcal{E} = \{\mathbf{e}_i\}_{ i= 1, \dots, n}$.

\subsection{External Medical KGs Retrieval}
To retrieve external medical knowledge relevant to the question, we query the KGs associated with each entity $\mathbf{e}_i$ in the set $\mathcal{E}$ from external knowledge base UMLS\footnote{The queried results are in the form of triplets, we use the term \textbf{KGs} instead of \textbf{triplets} to stress the knowledge structure.}~\cite{bodenreider2004unified}. The retrieved KGs defined as $G_i = (V_i, E_i)$, where $V_i$ represents a set of nodes (entities) and $E_i$ represents the relations between them. We integrate the extracted KGs $G_i$ to form the final external medical knowledge set, denoted as $\mathcal{G} = \{G_i\}_{ i=1, \dots, n}$.

\noindent\textbf{Caching Mechanism.} To accelerate retrieval, we construct a local knowledge base $B_{\text{local}}$ to store medical KGs retrieved from the remote UMLS. For each query $e_i$, we first search the local knowledge base. If the required medical entity is not found, we retrieve it from the remote UMLS and update $B_{\text{local}}$ accordingly. 

\begin{table*}[t]
\vspace{-3mm}
    \centering
    \resizebox{1.0\textwidth}{!}{
    \begin{tabular}{lllllllll}
        \toprule
         & \multicolumn{2}{c}{JMMLU} & \multicolumn{2}{c}{CMMLU} & \multicolumn{2}{c}{SW MMLU} & \multicolumn{2}{c}{KO MMLU} \\
        \cmidrule(lr){2-3} \cmidrule(lr){4-5} \cmidrule(lr){6-7} \cmidrule(lr){8-9}
         Model & Base & MKG-Rank & Base & MKG-Rank & Base & MKG-Rank & Base & MKG-Rank\\
        \midrule
         Qwen-2.5 72B~\cite{yang2024qwen2} & 74.00 & 80.22 (\textcolor[HTML]{009900}{+6.22\%}) & 84.54 & 81.60 (\textcolor[HTML]{CC0000}{-2.94\%}) & 44.28 & 50.90 (\textcolor[HTML]{009900}{+6.62\%}) & 67.72 & 71.86 (\textcolor[HTML]{009900}{+3.14\%})\\
         Llama-3.1 70B\textsuperscript{\textcolor{blue}{*}}~\cite{grattafiori2024llama} & 43.33 & 70.00 (\textcolor[HTML]{009900}{+26.67\%}) & 50.00 & 72.69 (\textcolor[HTML]{009900}{+22.69\%}) & 36.55 & 62.34 (\textcolor[HTML]{009900}{+25.79\%}) & 32.97 & 68.00 (\textcolor[HTML]{009900}{+35.03\%})\\
         Claude-3.5 haiku~\cite{claude35haiku} & 67.11 & 76.44 (\textcolor[HTML]{009900}{+9.33\%}) & 50.90 & 63.21 (\textcolor[HTML]{009900}{+12.31\%}) & 40.28 & 51.03 (\textcolor[HTML]{009900}{+10.75\%}) & 56.55 & 68.55 (\textcolor[HTML]{009900}{+12.00\%})\\
         GPT-4o-mini~\cite{gpt4omini} & 77.33 & 80.88 (\textcolor[HTML]{009900}{+3.55\%}) & 62.08 & 70.32 (\textcolor[HTML]{009900}{+8.24\%}) & 66.90 & 72.14 (\textcolor[HTML]{009900}{+5.24\%}) & 71.59 & 76.69 (\textcolor[HTML]{009900}{+5.10\%})\\
         GPT-4o~\cite{gpt4o} & 83.78 & \textbf{84.44} (\textcolor[HTML]{009900}{+0.66\%}) & 66.59 & \textbf{81.83} (\textcolor[HTML]{009900}{+15.24\%}) & 75.86 & \textbf{83.31} (\textcolor[HTML]{009900}{+7.45\%}) & 78.21 & \textbf{86.34} (\textcolor[HTML]{009900}{+8.13\%})\\
         \bottomrule
    \end{tabular}}
    \caption{Accuracy comparison between our proposed MKG-Rank and the base models on four multilingual datasets: JMMLU (Japanese), CMMLU (Chinese), SW MMLU (Swahili), and KO MMLU (Korean). \textcolor{blue}{*} indicates the base model on which MKG-Rank achieves the highest performance gain.  The best performance is shown in \textbf{bold}.}
    \label{tab:main_results}
    \vspace{-3mm}
\end{table*}

\subsection{Multi-Angle Ranking}

We propose Multi-Angle Ranking to extract the relevant medical knowledge, mitigating the impact of irrelevant or redundant information. Specifically, for the extracted medical knowledge set $\mathcal{G}$, we rank the medical triplets based on similarity scores with the question $T_q$, which are computed using UMLS-BERT~\cite{michalopoulos-etal-2021-umlsbert} embeddings. The re-ranked triplets, along with $T_q$ and $T_o$, are then processed through the MedCPT\footnote{\url{https://huggingface.co/ncbi/MedCPT-Cross-Encoder}} Cross Encoder for further filtering, finally obtaining the relevant medical knowledge set $\mathcal{G}^{'}$.



\subsection{Declarative Conversion}
\label{sec:deccomp}
For the relevant medical knowledge set $\mathcal{G}^{'}$, we use the LLM to convert it into declarative statements $\mathcal{S}^{'}$ in its preferred language. Additionally, since $\mathcal{G}^{'}$ may contain multilingual redundancy and some triples may lack sufficient useful information, we perform additional reasoning alongside declarative conversion to compress information that could negatively affect the LLM's reasoning. Finally, the question $T_q$ and options $T_o$, along with the medical knowledge in declarative form $\mathcal{S}^{'}$, are fed into the LLM for reasoning to generate the final answer, represented as:

\begin{equation*}
y = \text{LLM}(T_q, T_o,\mathcal{S}^{'}),
\end{equation*}

where $y$ represents the answer generated by the LLM, which uses the refined medical knowledge as supplementary information to enhance its reasoning in multilingual medical QA.

\subsection{Self-Information Mining}
In cases where the triplets retrieved from UMLS provide ineffective information, we employ the BM25~\footnote{\url{https://github.com/dorianbrown/rank_bm25}} algorithm for self-knowledge mining. This approach enables the model to extract relevant world knowledge from its own internal representations and retrieve the most pertinent fragments to enhance reasoning.


\section{Experiments}
\subsection{Datasets}
To evaluate the effectiveness of MKG-Rank in multilingual medical QA, we conducted experiments on four multiple-choice datasets covering different languages, focusing on medical-related subsets: JMMLU~\footnote{\url{https://huggingface.co/datasets/nlp-waseda/JMMLU}} (Japanese), CMMLU~\cite{li2023cmmlu} (Chinese), KO MMLU (Korean), and SW MMLU~\cite{singh2024global} (Swahili). Further details can be found in Appendix~\ref{appendix:dataset}.

\subsection{Evaluation Metric}
We use accuracy (Acc) as the evaluation metric, which measures the percentage of correct answers provided by the model. Furthermore, any response expressing uncertainty or listing multiple candidate answers is considered incorrect.


\subsection{Results and Analysis}

In Table~\ref{tab:main_results}, we compare MKG-Rank (LLM backbone) with the baseline LLMs (zero-shot setting). 
The results demonstrate that our method consistently outperforms all base LLMs. Specifically, with LLaMA 3.1 70B, we achieved over a 20\% improvement across all datasets. For large-scale closed-source LLMs, our method achieved the highest gain on Claude 3.5 Haiku, with an average improvement of 11.1\% across the four datasets. Additionally, we achieved an average improvement of 5.5\% and 7.8\% on GPT-4o-mini and GPT-4o, respectively.


Interestingly, we observed a performance drop for Qwen 2.5 72B on the CMMLU dataset. This is primarily because Qwen was trained on a large Chinese corpus, excelling in handling Chinese. As a result, integrating English medical knowledge instead interferes with its reasoning. As an extension, we conducted additional experiments on small-scale LLMs below 32B, as shown in Appendix~\ref{appendix:small}.

\begin{figure}[t]
    \centering
    \includegraphics[width=1\linewidth]{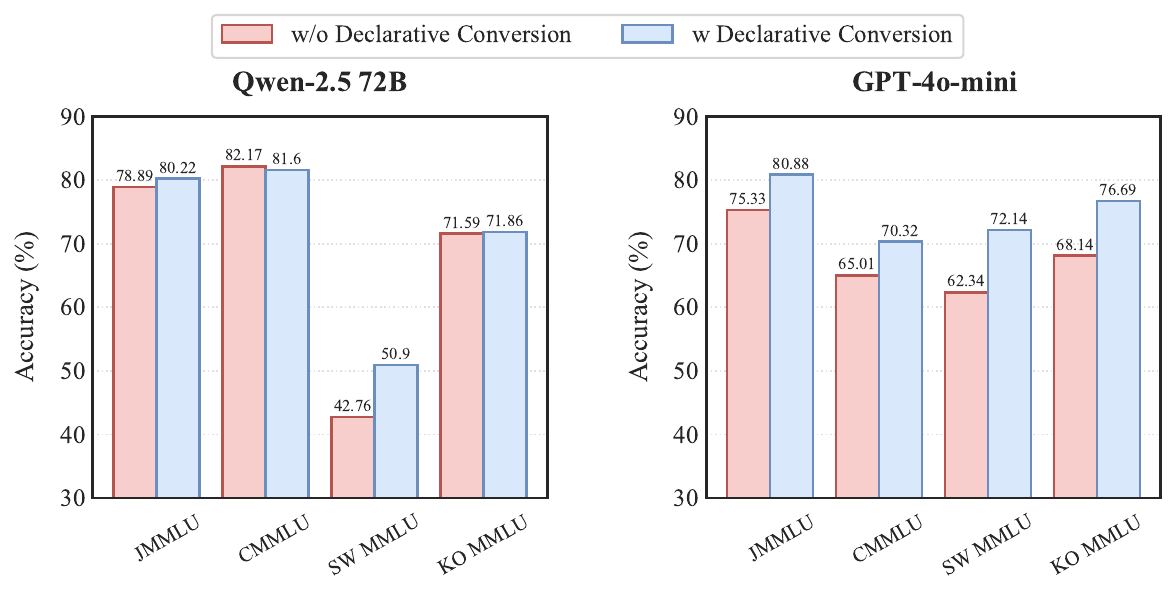}
    \caption{Comparison of the Acc evaluated on Qwen-2.5 72B and GPT-4o-mini across four language datasets with (w/) and without (w/o) declarative conversion.}
    \label{fig:ablation_rank}
    \vspace{-3mm}
\end{figure}

\subsection{Ablation Study and Analysis}

\noindent\textbf{Effectiveness of the Declarative Conversion.} 
To evaluate the effectiveness of the declarative conversion in Section~\ref{sec:deccomp}, we compare the performance of Qwen-2.5 70B and GPT-4o-mini with and without this mechanism, as shown in Figure~\ref{fig:ablation_rank}. The experimental results show that the declarative conversion mechanism significantly improves the performance of the base models, with particularly notable improvements observed on GPT-4o-mini. Directly retrieved knowledge graphs contain multilingual information, which can negatively affect the LLM’s encoding process, leading to accuracy degradation, especially on the SW MMLU dataset. The proposed declarative conversion mechanism addresses this issue by filtering and converting the retrieved raw data, allowing the model to focus on high-relevance English medical concepts while suppressing noise interference.


\noindent\textbf{Efficiency Analysis of Retrieval Optimization.} 
To evaluate the efficiency of our designed caching mechanism, we measured the average time for querying an entity with and without caching. Compared to the direct query time of \textbf{14} seconds, the caching mechanism reduces it to just \textbf{0.0009} seconds, achieving a speedup of four orders of magnitude. Additionally, the caching mechanism allows for real-time updates to the local knowledge base, providing a more effective solution for offline deployment.



\subsection{Case Study} 
\begin{figure}[t]
\vspace{-3mm}
    \centering
    \includegraphics[width=1\linewidth]{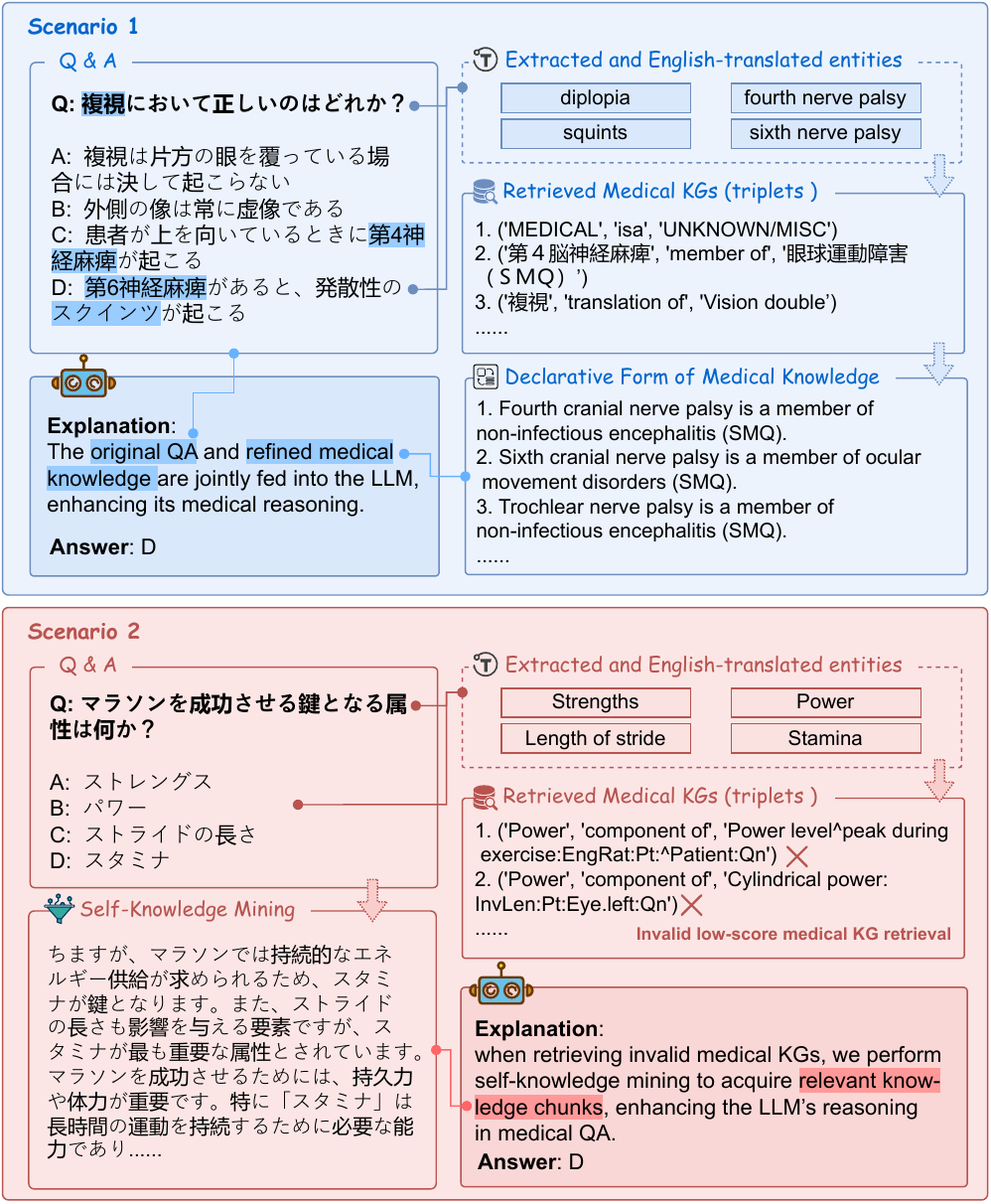}
    \caption{Case Study. More details, along with the English version of the questions and options are provided in the Appendix~\ref{appendix_case_study}.}
    \label{fig:case_study}
    \vspace{-5mm}
\end{figure}

We conduct case studies on two scenarios in our method, as shown in Figure~\ref{fig:case_study}. For scenario 1, we first extract relevant medical entities from the given medical question \& options and translate them into English (e.g., \textit{diplopia}, \textit{fourth nerve palsy}) for querying an external medical knowledge base. The retrieved medical KGs are multilingual and may contain redundant or irrelevant information. Our Multi-Angle Ranking strategy effectively filters out unrelated content. The filtered medical triples are then converted into statements in the preferred language of the LLM. Finally, the LLM makes the final decision by reasoning over the original question, options, and the obtained medical knowledge statements.

For scenario 2, we consider cases where the retrieved medical KGs are invalid (e.g., \textit{Cylindrical power}) due to the question and options containing minimal medical entities. In such cases, we filter out all irrelevant information, aiming to mine the model's knowledge to serve as background knowledge for the final reasoning process, thereby helping the LLM make the final decision.

\section{Conclusion}
In this work, we propose MKG-Rank, a knowledge graph-augmented framework that enables English-centric LLMs to effectively handle multilingual medical QA. By leveraging comprehensive external medical knowledge graphs and introducing a word-level translation mechanism, MKG-Rank effectively bridges the medical knowledge gap between English and other languages. Furthermore, we design Multi-angle ranking and caching strategies to enhance retrieval efficiency and filter relevant results, ensuring more accurate answers. Extensive experiments across four languages demonstrate that MKG-Rank consistently outperforms zero-shot LLM baselines in multilingual medical QA scenarios.

\section*{Limitations}
In this study, we developed an enhanced framework based on KG-Rank~\cite{yang2024kg} to improve the performance of LLMs in medical question answering. However, this framework also has certain limitations in practical applications, which we will discuss in the next phase. Firstly, for incremental databases, it is necessary to set a time for retrieval from the cloud to achieve a balance between efficiency and effectiveness. Secondly, for content that cannot be retrieved from the UMLS database, we only attempted BM25-based information mining, which is relatively simplistic. In the future, we plan to explore a reinforcement learning approach(~\cite{chen2025improving}) to strike a balance between exploitation and exploration, optimizing the reasoning process while leveraging the model's inherent knowledge.


\bibliography{custom}

\clearpage
\appendix
\onecolumn
\section{Dataset Details}
\label{appendix:dataset}
\noindent\textbf{JMMLU}~\footnote{\url{https://huggingface.co/datasets/nlp-waseda/JMMLU}}. It consists of a subset of Japanese-translated questions from MMLU~\cite{hendrycks2020measuring} and questions based on the Japanese cultural context. We selected three medically related subsets, which contain 450 entries.

\noindent\textbf{CMMLU}~\cite{li2023cmmlu}. It is a multi-task benchmark designed for Chinese language understanding, consisting of multiple-choice questions with four options.

\noindent\textbf{KO MMLU, SW MMLU}. Global MMLU~\cite{singh2024global} is a multilingual version derived from MMLU, which includes carefully translated and machine-translated versions in various languages. We select Korean and Swahili (a language widely spoken in East Africa) from this dataset, referred to as KO MMLU and SW MMLU, respectively.

From the aforementioned datasets, we select the medically related subsets, which include Clinical Knowledge, Professional Medicine, and College Medicine. More information is shown in Table~\ref{tab:dataset}.

\begin{table}[h]
  \centering
  \resizebox{0.42\textwidth}{!}{
  \begin{tabular}{lccc} 
    \toprule
    Dataset & Language & Size & Length \\
    \midrule
    JMMLU & Japanese & 450 & 171 \\
    CMMLU & Chinese & 886 & 70 \\
    SW MMLU & Korean & 725 & 511 \\
    KO MMLU & Swahili & 725 & 215 \\
    \bottomrule
  \end{tabular}}
  \caption{Statistics of evaluation datasets, including the size of each dataset and the average text length of each question and its corresponding options.}
  \label{tab:dataset}
\end{table}



\section{Resource Consumption}
\begin{table*}[h]
    \centering
    \resizebox{1.0\textwidth}{!}{
    \begin{tabular}{lcccccccc}
        \toprule
         & \multicolumn{2}{c}{JMMLU 450} & \multicolumn{2}{c}{CMMLU} & \multicolumn{2}{c}{SW MMLU} & \multicolumn{2}{c}{KO MMLU} \\
        \cmidrule(lr){2-3} \cmidrule(lr){4-5} \cmidrule(lr){6-7} \cmidrule(lr){8-9}
         Model & A100(hours) & API(\$) &A100(hours) & API(\$)  &A100(hours) & API(\$)  & A100(hours) & API(\$) \\
        \midrule
         Qwen-2.5 72B & 12 & 0.08 & 18 & 0.22  & 16 & 0.17 & 16 & 0.17 \\
         Llama-3.1 70B &  12 & 0.08 & 18 & 0.22  & 16 & 0.17 & 16 & 0.17 \\
         Claude-3.5 haiku& - & 1.84  & - & 3.6  & - & 2.97 & - & 2.97\\
         GPT-4o-mini & - & 0.26  & - & 0.54  & - & 0.44 & - & 0.44\\
         GPT-4o & - & 1.75  & - & 3.44  & - & 2.82  & - & 2.82\\
         \bottomrule
    \end{tabular}}
    \caption{Resources consumed in the relevant experiments.}
    \label{tab:main_consumption}
\end{table*}

\section{Additional Evaluation Experiments on Small-scale LLMs}
\label{appendix:small}
To further demonstrate the effectiveness of MKG-Rank, we evaluate its performance against small-scale baseline LLMs on the JMMLU dataset, as shown in Table~\ref{tab:appendix_experiment}. Experimental results show that MKG-Rank consistently outperforms all small-scale baseline LLMs.

\begin{table*}[h]
    \centering
    \resizebox{1.0\textwidth}{!}{
    \begin{tabular}{llllllll}
        \toprule
        Method & Borea-Phi-3.5 & Llama-3.2 3B & Qwen-2.5 7B & Meta-Llama-3.1 8B & Qwen-2.5 14B & Phi4 14B & Qwen-2.5 32B \\
        \midrule
         Base & 42.00 & 36.10 & 58.40 & 48.22 & 70.88 & 67.56 & 75.11 \\
         MKG-Rank & 43.43 (\textcolor[HTML]{009900}{+1.43\%}) & 39.78 (\textcolor[HTML]{009900}{+3.68\%}) & 61.33 (\textcolor[HTML]{009900}{+2.93\%}) & 52.22 (\textcolor[HTML]{009900}{+4.00\%}) & 73.11 (\textcolor[HTML]{009900}{+2.23\%}) & 79.56 (\textcolor[HTML]{009900}{+12.00\%}) & 76.00 (\textcolor[HTML]{009900}{+0.89\%})\\
         \bottomrule
    \end{tabular}}
    \caption{Accuracy comparison between MKG-Rank and small-scale base models on the JMMLU dataset.}
    \label{tab:appendix_experiment}
\end{table*}

\section{Additional Ablation Study on Declarative Conversion}
\begin{figure*}[h]
    \centering
    \includegraphics[width=0.75\linewidth]{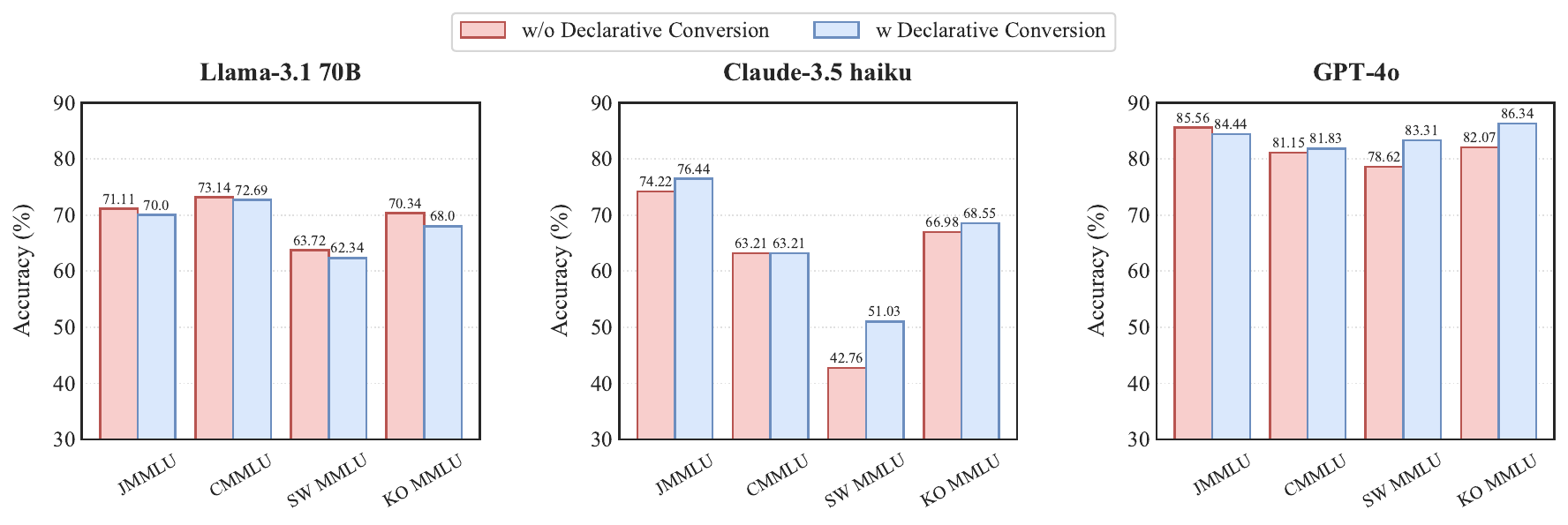}
    \caption{Additional ablation experiments on Llama 70B, Claude-3.5 haiku, and GPT-4o across four language datasets with (w) and without (w/o) multi-angle ranking.}
    \label{fig:appendix_ablation_rank}
\end{figure*}

We conducted an additional ablation study on Declarative Conversion across three LLMs, as shown in Figure~\ref{fig:appendix_ablation_rank}. Notably, on the LLaMA-3.1 70B model, Declarative Conversion demonstrates a negative impact, leading to a decline in performance. After analyzing the results, we believe that the longer transcription context, compared to directly using triples, imposes a greater reasoning burden than the multilingual effect. Specifically, the length of the context appears to affect the LLaMA model's performance more significantly.

\section{Prompts}
In this section, we will present the prompts used in each stage of reasoning within the MKG-Rank.
\label{sec:appendix}
\subsection{Medical NER Prompt}
\begin{figure*}[h]
    \centering
    \includegraphics[width=1\linewidth]{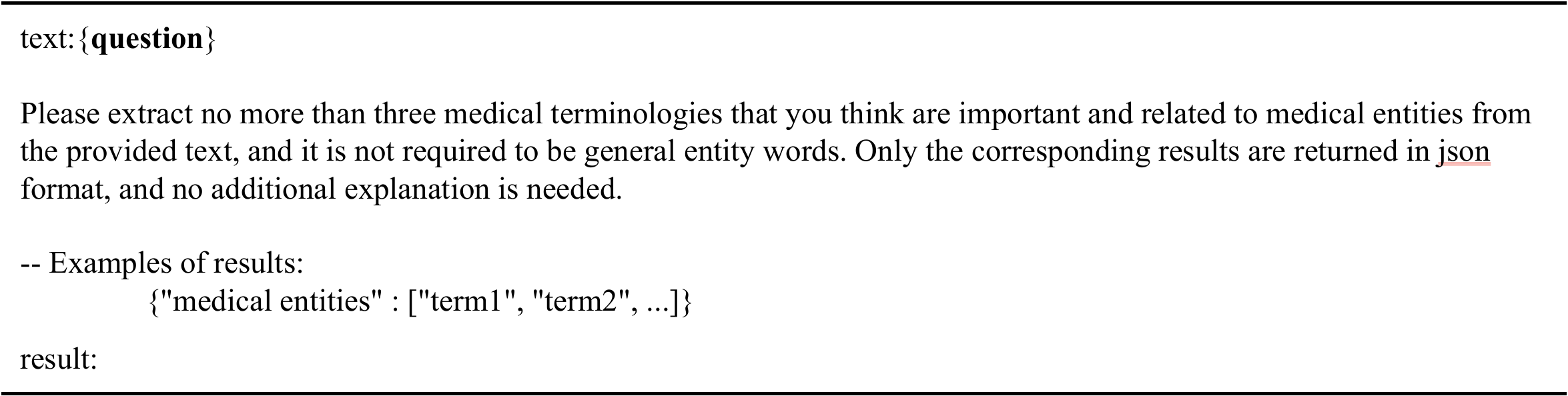}
    \caption{Prompts for extracting medical entities from question.}
    \label{fig:prompt_entity_extract}
\end{figure*}

\begin{figure*}[h]
    \centering
    \includegraphics[width=1\linewidth]{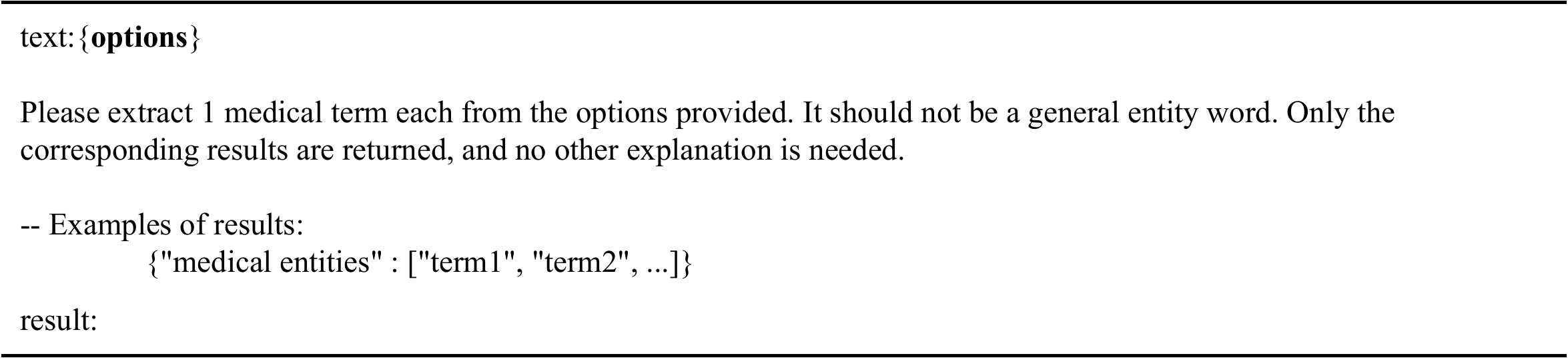}
    \caption{Prompts for extracting medical entities from options.}
    \label{fig:prompt_entity_extract2}
\end{figure*}

Figure~\ref{fig:prompt_entity_extract} and Figure~\ref{fig:prompt_entity_extract2} illustrate the prompt designed for extracting medical entities from both questions and options, with different extraction counts set for questions and options respectively.

\subsection{Declarative Conversion}
\begin{figure*}[h]
    \centering
    \includegraphics[width=1\linewidth]{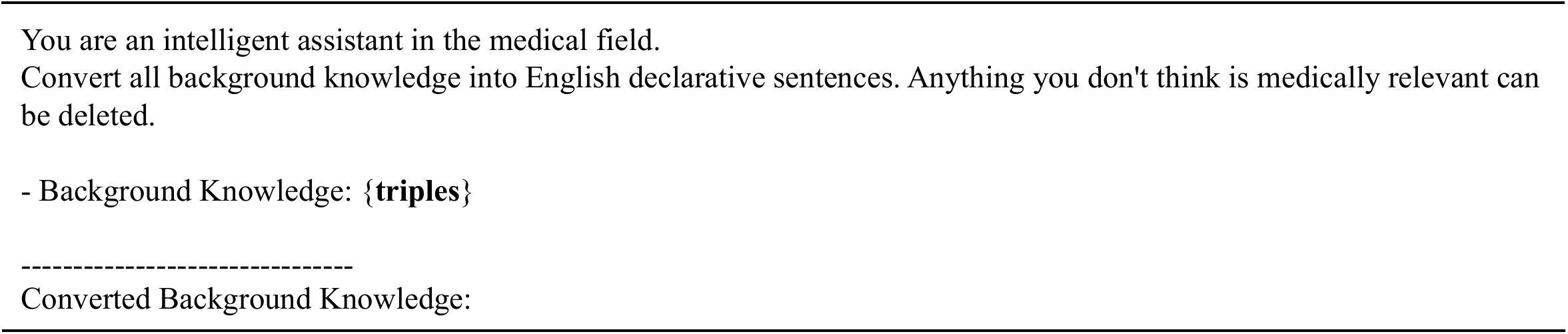}
    \caption{Prompts for declarative conversion.}
    \label{fig:convert_triplet}
\end{figure*}

Figure~\ref{fig:convert_triplet} illustrates the prompt designed for declarative conversion.

\subsection{MKG-Rank Enhanced Reasoning Prompt}
\begin{figure*}[h]
    \centering
    \includegraphics[width=1\linewidth]{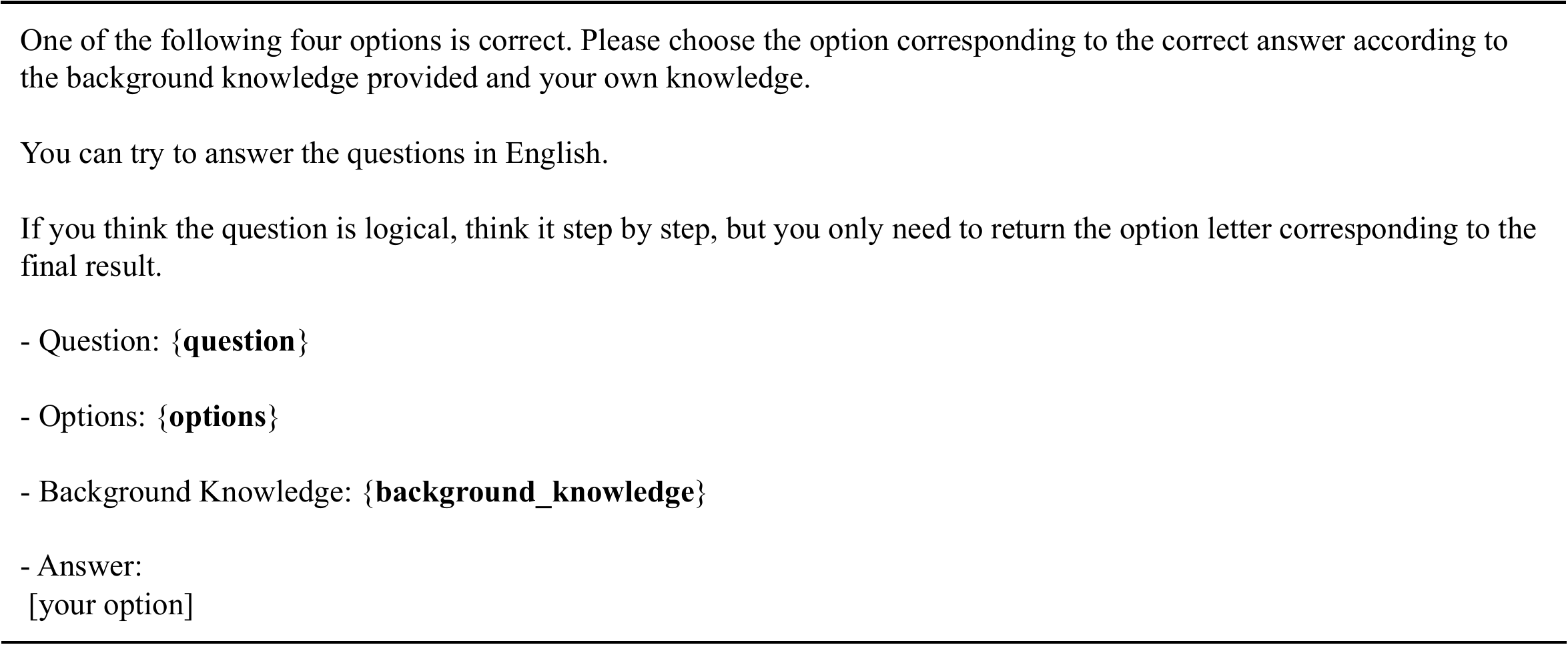}
    \caption{Prompts for MKG-Rank enhanced reasoning.}
    \label{fig:generate_answer}
\end{figure*}

Figure~\ref{fig:generate_answer} illustrates the prompt designed for reasoning based on the final integrated knowledge.
\\\\

\subsection{Self-Information Mining Prompt }
\begin{figure*}[h]
    \centering
    \includegraphics[width=1\linewidth]{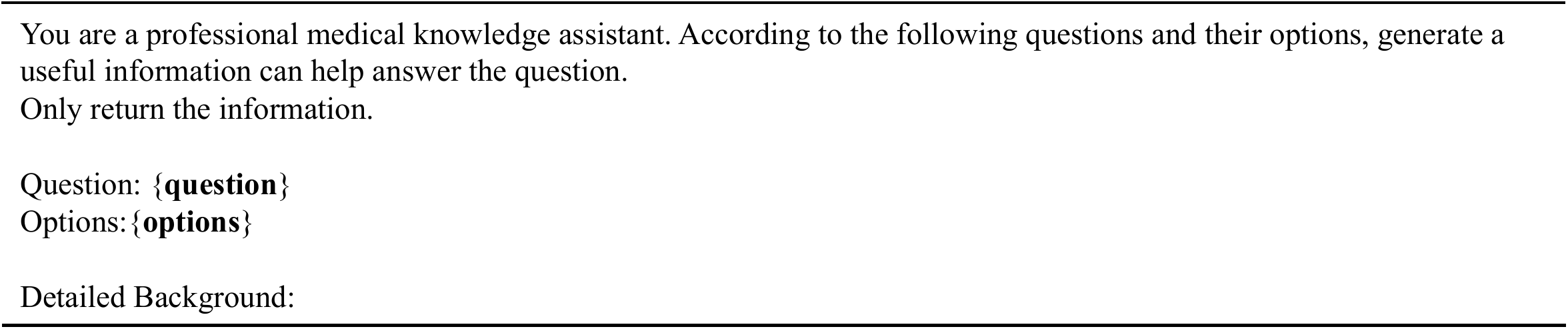}
    \caption{Prompts for Self-Information Mining.}
    \label{fig:information_mining}
\end{figure*}

Figure~\ref{fig:information_mining} illustrates the prompt designed to explore the model's intrinsic world knowledge, aiming to extract corresponding long-text information.

\section{English Translation of the Question and Options in 
Figure~\ref{fig:pipeline}}
\label{appendix_pipline}

\begin{figure*}[h]
    \centering
    \includegraphics[width=1.0\linewidth]{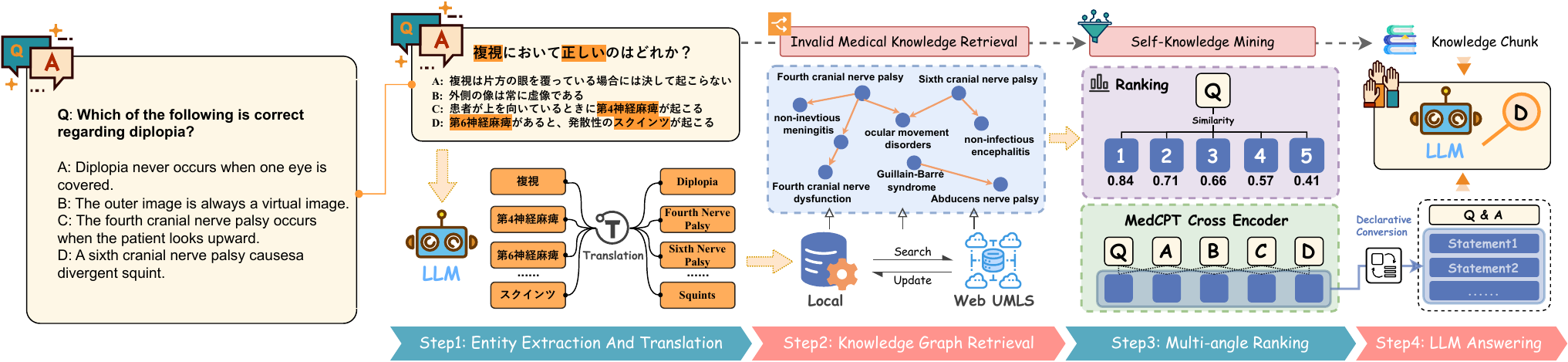}
    \caption{The English version of the question and options in Figure~\ref{fig:pipeline}.}
    \label{fig:english_pipline}
\end{figure*}

We provide the English translation of the question and options in Figure~\ref{fig:pipeline} for clearer description, as shown in Figure~\ref{fig:english_pipline}.

\section{A Detailed Case Study with English Annotations}
\label{appendix_case_study}
\begin{figure*}[h]
    \centering
    \includegraphics[width=0.66\linewidth]{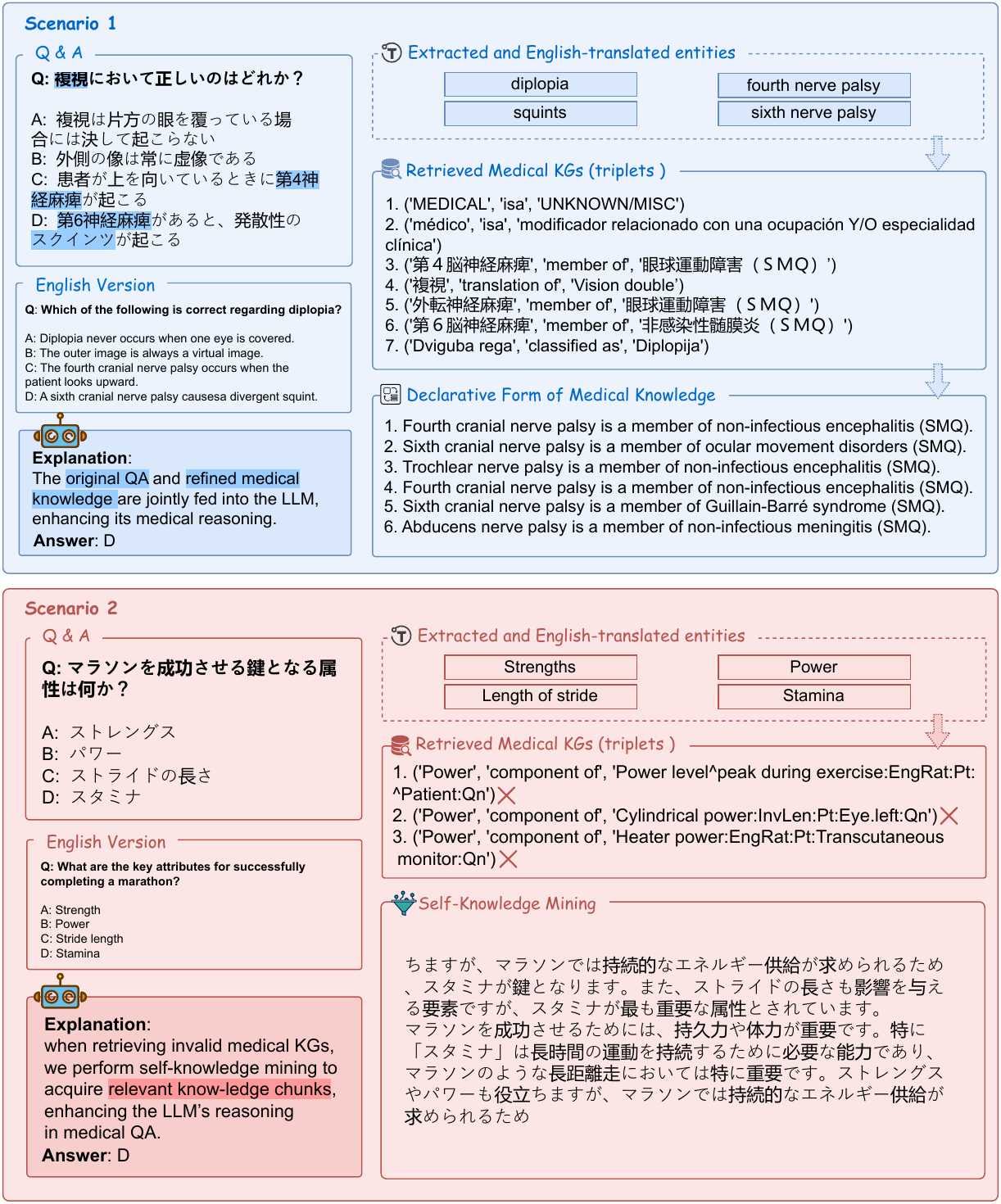}
    \caption{A detailed case study with comprehensive information, including the English version of the questions and options.}
    \label{fig:detail_cast_study}
\end{figure*}

We provide detailed information on the cases in Figure~\ref{fig:case_study}, along with the English version of the questions and options, as shown in Figure~\ref{fig:detail_cast_study}.

\end{document}